\documentclass{article}

\usepackage{arxiv}

\usepackage[utf8]{inputenc} 
\usepackage[T1]{fontenc}    
\usepackage{hyperref}       
\usepackage{url}            
\usepackage{booktabs}       
\usepackage{amsfonts}       
\usepackage{nicefrac}       
\usepackage{microtype}      
\usepackage{lipsum}

\usepackage[
    backend=biber,
    style=ieee,
  ]{biblatex}

\usepackage{graphicx}  
\usepackage{subcaption}
\usepackage{booktabs}
\usepackage{mathtools}
\usepackage[flushleft]{threeparttable}
\usepackage{url}
\usepackage{authblk}
\usepackage{floatrow}
\usepackage{cleveref}
\usepackage{dashrule}
\usepackage{longtable}

\graphicspath{ {./figs/} }
    
\title{Characterizing Partisan Political Narrative Frameworks about COVID-19 on Twitter}
\author[\space\space 1]{Elise Jing\thanks{Current address: Sirius XM, 1221 Avenue of the Americas 37th Floor, New York, NY 10020}}
\author[123]{Yong-Yeol Ahn}
\affil[1]{Center for Complex Networks and Systems Research, Luddy School of Informatics, Computing and Engineering, Indiana University Bloomington, IN 47408, USA}
\affil[2]{Network Science Institute, Indiana University, Bloomington (IUNI), IN 47408, USA}
\affil[3]{Connection Science, Massachusetts Institute of Technology, Cambridge, MA 02139, USA}

\begin{document}

\maketitle
\begin{abstract}
The COVID-19 pandemic is a global crisis that has been testing every society and exposing the critical role of local politics in crisis response. In the United States, there has been a strong partisan divide between the Democratic and Republican party's narratives about the pandemic which resulted in polarization of individual behaviors and divergent policy adoption across regions. As shown in this case, as well as in most major social issues, strongly polarized narrative frameworks facilitate such narratives. To understand polarization and other social chasms, it is critical to dissect these diverging narratives. Here, taking the Democratic and Republican political social media posts about the pandemic as a case study, we demonstrate that a combination of computational methods can provide useful insights into the different contexts, framing, and characters and relationships that construct their narrative frameworks which individual posts source from.  Leveraging a dataset of tweets from the politicians in the U.S., including the ex-president, members of Congress, and state governors, we found that the Democrats' narrative tends to be more concerned with the pandemic as well as financial and social support, while the Republicans discuss more about other political entities such as China. We then perform an automatic framing analysis to characterize the ways in which they frame their narratives, where we found that the Democrats emphasize the government's role in responding to the pandemic, and the Republicans emphasize the roles of individuals and support for small businesses. Finally, we present a semantic role analysis that uncovers the important characters and relationships in their narratives as well as how they facilitate a membership categorization process. Our findings concretely expose the gaps in the ``elusive consensus'' between the two parties. Our methodologies may be applied to computationally study narratives in various domains.
    \end{abstract}


\section{Introduction}\label{sec:introduction} 
Human beings make sense of the reality around them by constructing narratives using what they see, hear, and encounter~\cite{patterson1998narrative}. However, narratives that evolve around different identities, cultures, religions, etc. are often at odds with each other~\cite{moon2012and}. One of the areas where contrasting narratives fiercely collide and fight is politics. Political communication often happens through narratives and stories, rather than logical reasoning~\cite{bruner2009actual, haidt2007moral}. These narratives have a tremendous power in shaping people's stances and behaviors on important social issues~\cite{kubin2021personal}. In the age of social media, narratives can be circulated, mutated, and amplified with incredible intensity and speed~\cite{bessi2015science, fan2020crowd}. For example, during the COVID-19 crisis, social media sites including Twitter and Facebook are used by the the anti-mask and anti-vaccine groups to organize multiple anti-mask protests~\cite{grimes2020health}. The anti-mask and anti-vaccine narratives, accompanied by conspiracy theories, fake news, and unverified anecdotes, discouraged mask usage and vaccination heavily, which might have led to the loss of hundreds of thousands more lives~\cite{ihme2020modeling}. Furthermore, such narratives often lead to collisions between partisan beliefs that strengthens political polarization~\cite{schmidt2018polarization}. As can be seen in the case of the pandemic narratives, understanding social conflicts and polarization is often impossible without understanding diverging narratives.

While many different definitions of narratives have been proposed, here we draw our definition of political narratives from the Narrative Policy Framework which defines a narrative as having ``(i) a setting or context; (ii) a plot that introduces a temporal element, providing both the relationships between the setting and characters, and structuring causal mechanisms; (iii) characters who are fixers of the problem (heroes), causers of the problem (villains), or victims (those harmed by the problem); and (iv) the moral of the story''~\cite{jones2010narrative}. Using this definition, we can identify narratives used by politicians and political parties to convey their morals. For example, as summarized by~\textcite{haidt2009above}, the liberals and conservatives in the United States have the following constrasting narratives that their followers adopt: \emph{``The majority of people used to be oppressed, treated unequally and with unjust; however, the courageous people fought against the powers and freed a lot of the oppressed people. We as successors must continue their errand and fight for more equality in the society.'' } (the ``liberal progress'' narrative from the liberals) Or,

\emph{``People used to live in harmonious communities tied together by faith and tradition, however, this is broken by the modern lifestyle, science and the industrial revolutions. We must therefore hold to our values and resist these forces.''} (the ``community lost'' narrative from the conservatives)

These narratives are not objective descriptions of history, but interpretations of the reality that fit with people's political beliefs. Additionally, even though the narratives are different and may be at conflict with each other, each of them achieve internal consistency and coherence~\cite{smith1989narrative}, which makes them effective~\cite{fisher1989human}. 

Traditional studies of political narratives are often based on political discourse analysis (PDA). PDA studies the role of spoken and written language in politics~\cite{dunmire2012political}, focusing on the rhetoric features, styles, logic, metaphors, and contents of the political language~\cite{charteris2018analysing}. While traditional PDA often draws its material from formal political language such as public speeches from national leaders~\cite{charteris2004angel}, legislative debates~\cite{park2014analyzing}, and newspaper articles~\cite{garretson20088}, social media has gained increasing attention as many politicians turn to social media sites as their main online platforms for public communication~\cite{enli2017twitter}, where they respond to issues raised by the media and public and promote their own agendas~\cite{barbera2019leads}.

Among social media sites, Twitter has been one of the most important platform for political discourse during the last decade~\cite{parmelee2011politics}. Politicians use Twitter to not only broadcast to, but also interact with and attract their audience directly~\cite{doi:10.1080/1369118X.2013.782330, doi:10.1080/15295036.2016.1266686}. Such direct communication often benefits politicians; for instance, the usage of Twitter may increase the amount of donation that a politician receives and benefit their campaigns~\cite{hong2013benefits, lee2012they}. For these reasons, as well as the succinct, swift, and amplifying nature of the Twitter discourse, many politicians have been effectively using their tweets to spread their narratives~\cite{johnson2016all}. While there have been studies on the hashtags~\cite{hemphill2013framing}, sentiments~\cite{kouloumpis2011twitter}, and moral values~\cite{johnson2018classification} from the politicians' tweets, systematic studies of political narratives on Twitter are rare, although political science increasingly adopts text analysis methods~\cite{wilkerson2017large}.

While the scale of social media data provides great opportunities, it also poses many challenges. Traditional approaches to narrative studies through ``close reading''~\cite{moretti2000conjectures} may allow deep understanding of narratives, but are labor-intensive and rely on subjective judgements. Such constraints may be addressed by computational methods, where we can automatically identify patterns in large datasets. For example, \textcite{shurafa2020political} studied hashtags and rhetoric devices used by U.S. Twitter users leaning towards the Democratic or Republican parties, and identified their framing preference regarding the COVID-19 crisis;  \textcite{green2020elusive} identified key words from politicians' tweets, and showed that partisanship can be inferred by their word usage. However, these studies rely on word-level analysis and Twitter hashtags, while in-depth analysis of such narratives are rarely attempted. 

Additionally, the brief nature of Twitter postings makes it unlikely for each of them to contain a complete narrative. Rather, each tweet may contain ``fragments'' of a larger narrative. While human readers can often infer the overarching narrative based on their reading of other tweets and background knowledge, it is difficult for computational models to do so. A similar challenge is identified by Tangherlini, et al. in their study of online conspiracy theories~\cite{tangherlini2020automated}, where the complete narrative is often scattered in multiple short postings. Their response is to consider a narrative framework consisting of ``cast of characters, the relationships between those characters, the contexts in which those relationships arise'', which individual postings sample from. Similarly, we consider two narrative frameworks for the U.S. Democratic and Republican parties, which are conceptualized by the aggregation of each party's tweets respectively, containing the contexts, characters, and relationships used by each party's narrative. Individual tweets draw their ``ingredients'' from this larger space, and allude to the complete narrative therein.

Following this intuition, we characterize the narrative frameworks for the two parties by analyzing collections of their tweets to identify three elements: context, framing, and characters and relationships. Our approach has two key differences from ~\textcite{tangherlini2020automated} in that (i) we consider the context as the main topics and issues that each party engages with, instead of characterizing it with relationships. (ii) we examine framing separately as we consider it to be a central piece of political discourse, which shapes \emph{how} political narratives are conveyed to the audience independent from \emph{what} is communicated (we further elaborate on this below). In doing so, we aim to provide more nuanced analysis beyond the common term-based approaches. 

First, we analyze the word frequencies in the tweets and identify the most characteristic words used by each party; this simple method allows us to see the most contrasting differences in each group's narratives at the level of ``ingredients'', which set up the contexts for their narrative frameworks. 

Next, we ask how they are \emph{framed}. Framing analysis is a central piece in political discourse analysis~\cite{wang2016new}. Framing is about selectively presenting some aspects of an issue and make them more salient, in order to promote certain values, interpretations, or solutions~\cite{entman1993framing}.  For example, on the undocumented immigration issue, the Democrats often focus on the human rights aspect, while the Republicans often focus on the legality. Similar divergence in framing across major political issues are widely recognized from the two parties. \textcite{hemphill2013framing} showed that using Twitter, a machine learning classifier can be trained to easily predict the partisanship of a politician from the frames that they use.

Traditional studies on political framing mostly rely on manual content analysis and discourse analysis to detect frames from texts~\cite{pan1993framing}, and are therefore confined to a small set of frames because the process is labor-intensive. Here, we employ the FrameAxis model~\cite{kwak2020frameaxis}, which was developed to facilitate this process by using word embeddings and antonymous word pairs. With this method, the overall \emph{bias} (the alignment with a frame) and \emph{intensity} (the strength of a frame) of a document with respect to many ``microframes'' can be computed. We apply the FrameAxis to identify important frames in the politicians' tweets about COVID-19. For example, we found the microframe \emph{dead vs. live} is used to discuss the deaths related to COVID-19, and the microframe \emph{fast vs. slow} is used to discuss the spread of COVID-19.

Finally, we analyze the characters and relationships in each party's narrative framework. We focus on the relationships captured by actions, the Agent (the one who initiates an action), and the Patient (the one being affected or the recipient of the action). For example, in the sentence \emph{Joe Biden was elected as president.}, \emph{Joe Biden} is the Agent, \emph{president} is the patient, and the relationship is captured in the verb \emph{elect}. The Agent--Patient--Action pattern appears to be universal in human cognition~\cite{cohn2013prediction}. 

We use semantic role labeling (SRL) models to automatically identify Agents, Patients, and verbs in our dataset. Originated in traditional linguistics~\cite{fillmore1967case}, SRL has attracted much interest from Computational Linguistics, leading to the development of large annotated corpora such as FrameNet~\cite{baker1998berkeley} and PropBank~\cite{kingsbury2002treebank}. Trained on such corpora, modern NLP platforms such as SENNA and AllenNLP can perform the SRL task with high accuracy~\cite{collobert:2011b,Gardner2017AllenNLP}. With the development of deep learning, SRL has been successfully applied to analyze events either as a stand-alone work or as part of an NLP pipeline~\cite{hung2010web, exner2011using, llorens2013applying}. As different semantic roles can refer to the same underlying character (e.g. ``Kamala Harris'' and ``Vice President Harris'' refer to the same person), other NLP techniques such as named entity recognition and coreference resolution are sometimes used to aggregate similar semantic roles and verbs~\cite{tangherlini2020automated,ash2021text}.

We are especially interested in the characters that play key roles in the COVID-19 crisis and the relationships between them. For example, when the Democrats use the word ``help'', \emph{who are to be helped and who will help them?} Furthermore, how are these agents different in the Republican tweets? Our analysis shows the most prominent Agents and Patients in the Democratic/Republican narratives about the pandemic as well as the partisan differences. In particular, we identify a membership categorization process, namely the division between ``us'' and ``them''. As the most general membership categories, they help people to organize their everyday knowledge and actions~\cite{sacks1992lectures}. For example, the former President Donald Trump frequently used this categorization in his campaign: ``They hate me. They hate you. They hate rallies and it's all because they hate the idea of MAKING AMERICA GREAT AGAIN!''~\cite{trumpusthem}. Our analysis reveals a similar process where memberships are established by the interaction between characters.

Overall, our work applies a set of computational methods to comprehensively describe the elements making up the two parties' narrative frameworks, as well as how they diverge. Such divergence may be one of the ``wedges'' that exacerbate polarization in U.S. politics. The combination of methods we employed here to explore political narratives are not limited to politics. The code we develop and publish would allow similar automatic analysis in various domains.

\section{Data and Methods}
\label{sec:data} 

\begin{table}[h]
\small
\begin{center}
\begin{tabular}{ccc} 
\hline
 & Number of tweets & Average number of tweets per politician\\
\hline
Senate (Republican)   & 34,329 &635.7 \\
Senate (Democratic) & 38,539 & \textbf{820.0}\\
\hline
House (Republican) & 108,095 & 420.6\\
House (Democratic) &205,746 & \textbf{635.2}\\
\hline
Governor (Republican) & 23,397 & 899.9\\
Governor (Democratic) & 30,398 & \textbf{1085.6}\\
\hline
Former President Trump & 1,196 & 1,196\\
\hline

Total (Republican) & 167,017 & 494.1\\
Total (Democratic) & 274,683 & \textbf{704.3}\\

\hline
\end{tabular}
\caption{The number of tweets posted by each group of politicians and the average number of tweets posted per person.} 
\label{table:tweet_size}
\end{center}
\end{table}

We collect data from major U.S. politicians on Twitter. Using the Twitter lists created by \texttt{cspan}\footnote{https://twitter.com/cspan}, we retrieve screen names of politicians including: U.S. Senators, House Representatives, state governors, and former President Trump. These Twitter accounts may be managed by the politicians or their staff, but in either case, they convey the messages from these politicians and are integral parts of their public images. We collect tweets from these accounts monthly starting in April 2020. In this study, we use tweets timestamped between February 1, 2020---one week after Wuhan's lockdown started---to July 22, 2020. We use the full texts of tweets and only keep the English tweets. 

\begin{figure}[h]
    \centering
\includegraphics[width=\textwidth]{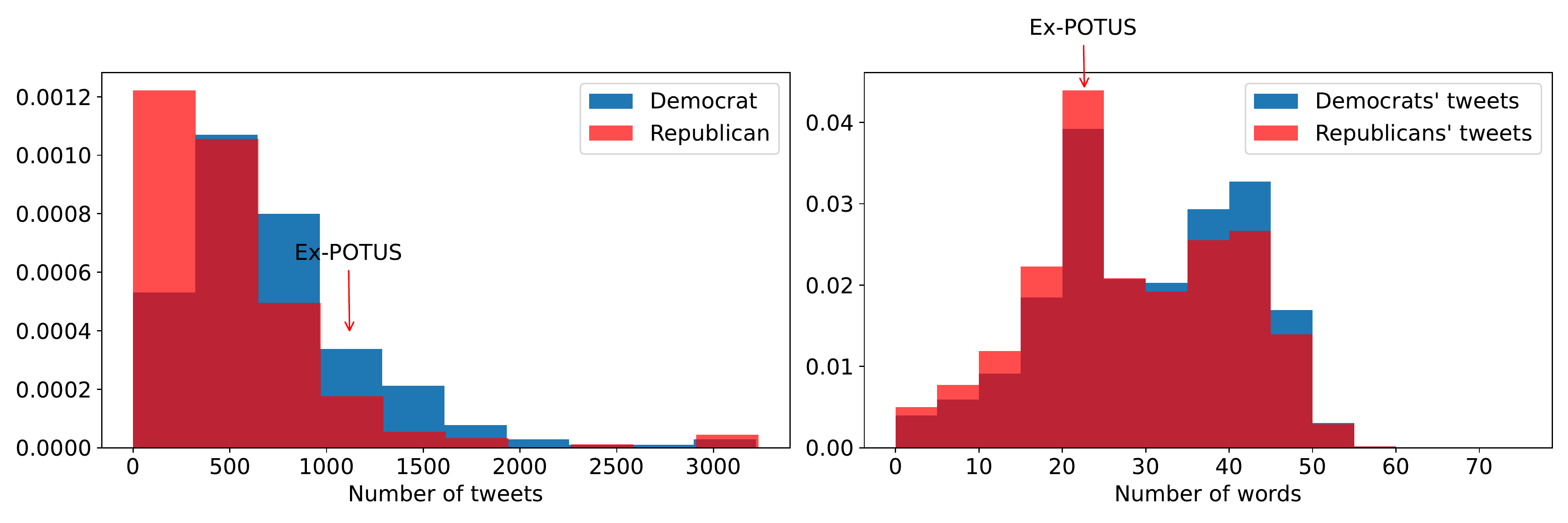}
    \caption{The distribution of the amounts of tweets that politicians post (left) and the length distribution of tweets (right). }
    \label{fig:dist}
\end{figure}

The number of politicians' tweets from each group is summarized in Table \ref{table:tweet_size}. We found that the Democratic politicians tend to post more compared to their Republican peers. Figure \ref{fig:dist} shows the distribution of politicians' posting frequencies and the length distribution of the tweets. We found a highly skewed distribution, where a few politicians tweet often and most only tweet occasionally. The majority of tweets have between 20--50 words for both groups.


\subsection{Filtering COVID-19 related tweets}
Because we are most interested in the COVID-19 related political discourse, we identify COVID-19 related tweets by checking if ``COVID'' or ``coronavirus'' is present in a tweet (case insensitive). This may omit some tweets that are about the pandemic but do not mention the name, but it ensures that all tweets we consider are related to COVID-19. The number of COVID-19 related versus non-related tweets are show in \Cref{table:covid_tweet}.

\begin{table}[h]
\begin{center}
\begin{tabular}{ccc} 
\hline
 & Number of COVID-19 related tweets & Number of non-related tweets\\
 \hline
Republican & 37,854 & 127,275\\
Democratic & 61,944 & 212,050\\
\hline
\end{tabular}
\caption{The number of COVID-19 related tweets and non-related tweets for each party}
\label{table:covid_tweet}
\end{center}
\end{table}

\subsection{Identifying over-represented terms}
For an overall understanding of the topics and key issues that set up the contexts of each party's narrative framework, we identify the over-represented words in their tweets. We use the log-odds ratios with informative Dirichlet priors~\cite{monroe2008fightin} by computing the log-odds ratio of each word $w$ in two corpora $i$ and $j$, with a background corpus $bg$ as prior. This is formally expressed as:
\begin{equation}
s_w = \log\frac{f_i+f_{bg}}{n_i + n_{bg} - f_i + f_{bg}} 
- \log\frac{f_j + f_{bg}}{n_j + n_{bg} - f_j + f_{bg}}
\end{equation}

where $f_i$ is the frequency of the word in the target corpus; for example, words in the COVID-19 related Democratic tweets. $f_{bg}$ is the frequency of the word in the background corpus. In this case, it is the combination of the Democratic and Republican tweets that are not related to COVID-19. $n_i$ is the size of the target corpus, and $n_{bg}$ is the size of the background corpus. $f_j$ is the frequency of the word in the \emph{other} corpus, in this case, the COVID-19 related Republican tweets; and $n_j$ is the size of this corpus.

Furthermore, we compute the $z$-scores of the log odds ratio as:
\begin{equation}
z_w = \frac{s_w}{\sqrt{\frac{1}{f_i+f_{bg}} + \frac{1}{f_j+f_{bg}}}}
\end{equation}
where the denominator serves as an estimate of the variance of the log-odds ratio.

We choose the top 40 words with highest $z$-scores from each party's COVID-related tweets as the most over-represented words. We exclude the politicians' names and Twitter handles as tend to be over-represented in each party's tweets. To better explore these words and the topics they represent, we obtain their contextual embeddings using word embedding models. While many word embedding models are available, we choose the GloVe~\cite{pennington2014glove} embeddings as it is considered one of the most effective word embedding models~\cite{naili2017comparative} and is widely used.

As many of the topic words are specific to the COVID-19 crisis, we train a new GloVe model on our tweet corpus for 500 epochs\footnote{Training for less epochs result in less distinct clustering of the embeddings, but does not change the overall result.} to obtain embeddings for words not in the pre-trained GloVe models. We use word vectors with a dimensionality of 300. Furthermore, for a consistent representation for terms related to ``COVID'', we compile a list of all tokens including ``COVID'' or ``coronavirus'' and replace them with ``COVID'' in the corpora. 

To explore the topic words visually, we use the Uniform Manifold Approximation and Projection (UMAP), an effective~\cite{yang2021dimensionality} and efficient~\cite{becht2018evaluation} dimensionality reduction method, to reduce the dimensionality of the GloVe embeddings. This method works by finding low-dimensional projections of the data that preserves their topological structures in high-dimensional space as much as possible~\cite{mcinnes2018umap}. We use the Python package \texttt{umap}. We plot the word embeddings with the dimensionality reduced to 2. With this visual aid, we identify and manually label six clusters for the Democratic tweets and three for the Republican tweets (see Section~\ref{sec:results}).

\subsection{Microframe Analysis}
Most of the traditional framing analysis methods rely on ``close reading'' and manual examination of linguistic material, and are therefore challenging to apply  to our dataset. Here, we employ the FrameAxis model~\cite{kwak2020frameaxis}, which allows an exploratory framing analysis through ``microframes''. A microframe is operationalized as a pair of antonyms, such as ``legal'' and ``illegal'', or ``fast'' and ``slow''. In political science research, usage of antonyms has been successfully capturing political stances. For example, the Moral Foundations Theory uses five pairs of antonyms such as ``Care/Harm'' and ``Fairness/Cheating'' to serve as moral ``axes''~\cite{haidt2007morality}. Here we use 1,621 antonym pairs obtained from WordNet~\cite{miller1995wordnet}. 

We then compute the \emph{bias} and \emph{intensity} of each microframe present in a document based on the vector representations of the microframes and other words in the text. We define the contribution of a word to a microframe as the cosine distance between the word vector $w$ and the microframe's vector $f$ (see \textcite{kwak2020frameaxis} for details):
\begin{equation}
c^w_f = \frac {v_w \cdot v_f}{\parallel{v_w}\parallel \parallel{v_f}\parallel}.    
\label{eq:cosine_similarity}
\end{equation}

The bias of a microframe is defined as the average contribution of all words in the document to the microframe. It captures the stance of a political argument; for example, 
a conservative document on the immigration issue may be biased towards \emph{illegal} rather than \emph{legal} in the \emph{illegal versus legal} microframe. Formally, the bias is computed as

\begin{equation}
    \mathrm{B}^t_f = \frac{\sum_{w \in t} (n_w  c^w_f) }{\sum_{w \in t} n_w},
\label{eq:frame_bias}
\end{equation}
where $t$ is a document, $f$ is a microframe, and $n_w$ is the number of occurrences of word $w$ in $t$.

Meanwhile, the intensity of a microframe captures how strongly it is presented in a document, regardless of which ``pole'' the document is closer to. The intensity is computed using the second moment of the word contribution with a background corpus as baseline:
\begin{equation}
    \mathrm{I}^t_f = \frac{\sum_{w \in t} n_w (c^w_f - \mathrm{B}^T_f)^2}{\sum_{w \in t} n_w},
\end{equation}
where $\mathrm{B}^T_f$ is the baseline {microframe} bias of the entire text corpus $T$ on a microframe $f$ for computing the second moment. As the squared term is included in the equation, the  words that are far from the baseline {microframe} bias---and close to either of the poles---contribute strongly to the {microframe} intensity.

Here we compute the bias and intensity for each COVID-19 related tweet, using a background of non-COVID-19 related tweets, for each microframe. We focus on the microframes with the largest difference in intensity between the two parties; for the Democratic party, we present the microframes where the intensity in Democratic tweets is higher than that in the Republican tweets, and vice versa. In addition to showing the microframes, we also show the top 3 tweets with the strongest intensity for each microframe.

\subsection{Semantic Role Analysis}
To identify important semantic roles, we use the Python package \texttt{Allennlp}~\cite{Gardner2017AllenNLP} to perform semantic role labeling on our corpus. We focus on the verb, the Agents (\emph{Arg0} in the \texttt{Allennlp} system), and the Patents (\emph{Arg1}). To focus on the most common semantic roles, we only consider the Agents and Patients consisting of three or less tokens.

 To obtain a list of semantic roles specifically related to the Democratic and the Republican party, we produce two lists of terms most similar to the words ``Democrat'', ``Democratic'', and ``Republican'' using the fine-tuned GloVe embeddings we described above. The terms most similar to ``Democrat'' and ``Democratic'' include ``dems'', ``housedemocrats'', ``reddemocrats'', ``democraticled'', ``pelosi'', ``speakerpelosi'', ``nancy pelosi'',  ``chuck schumer'', ``ralph northam'', ``ayanna pressley'', ``gwen moore'', and ``senatedems''. The terms most similar to ``Republican'' include ``gop'', ``republicans'', ``president'', ``trump'', ``donald trump'', `patrick mchenry'', ``larry hogan'', ``mitch mcconnell'', and ``mcconnell' (case insensitive).

We identify important verbs by considering the top 100 most frequent verbs in each party's tweets. We obtain the GloVe embeddings for each verb in the same manner as we describe above. We then use UMAP to reduce the dimensionality of the embeddings, and then use the k-means clustering algorithm to group the verbs from each party into 15 clusters. This produces clusters of verbs that are semantically close to each other in daily usage, but also indicates some verb usage that are specific to parliamentary politics.

\section{Results}\label{sec:results} 

\begin{figure}
\centering
\includegraphics[width=\textwidth]{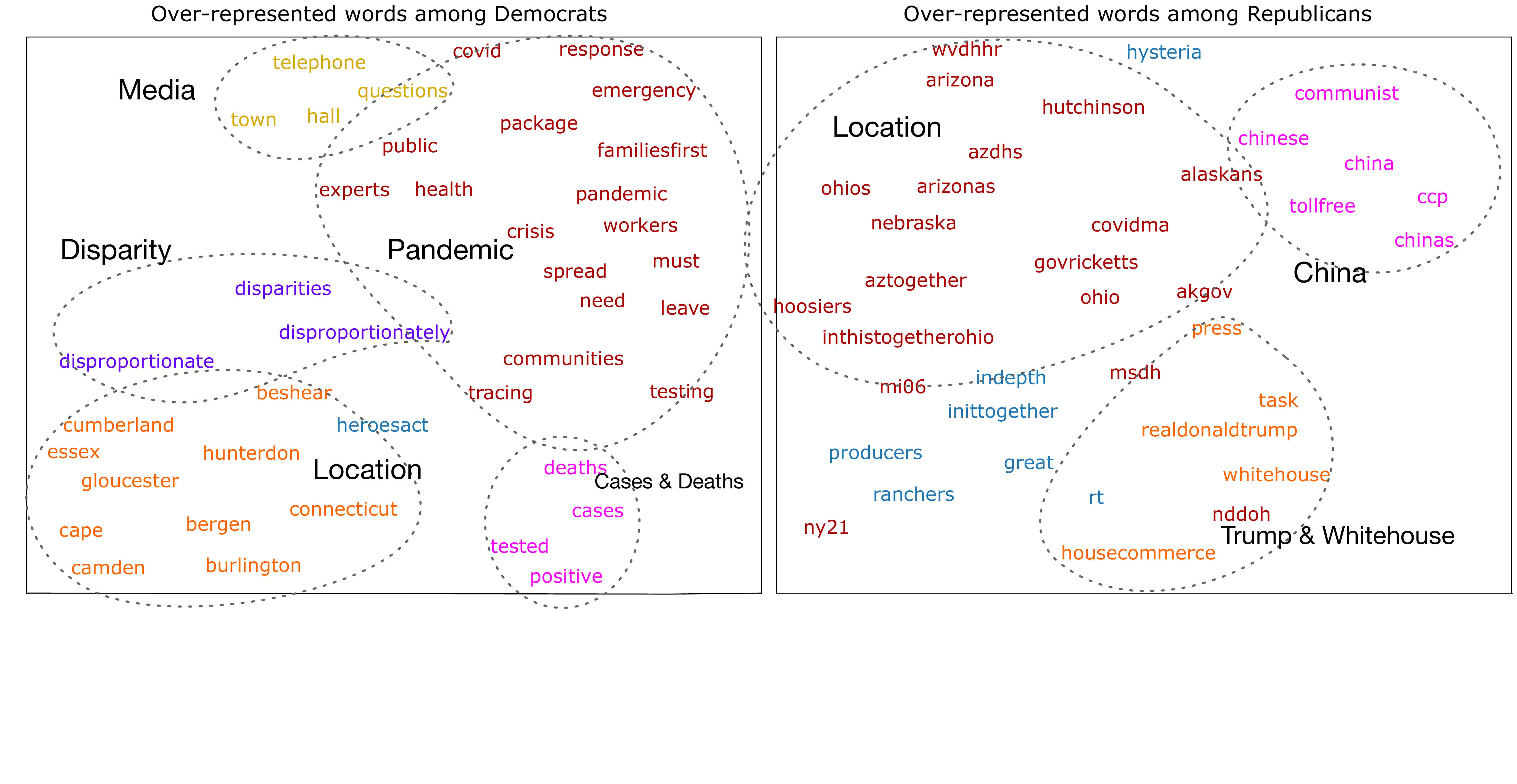}
\caption{Characteristic words in each party's tweets related to COVID-19 in the GloVe word embedding space. We detect over-represented words by calculating the log odds ratio of each word (see Section~\ref{sec:data}) and obtain the GloVe embeddings for each word. We use UMAP to reduce dimensionality and plot each word. Colors indicate topic labels that we assign. The Democratic party member's tweets features more words about the pandemic and about the pandemic and its disproportionate influences, while the Republican tweets features words about Trump and the White House as well as words about China.	
}
\label{fig:logodds}

\end{figure}

First, we look at the most characteristic words found in each party's tweets. We start with comparing each word's dense rank~\cite{kessler2017scattertext} in the COVID-related Democratic and Republican tweets and the background corpus to find words over-represented in the COVID-related tweets. While these tweets unsurprisingly features many shared words between parties as shown in Table S1, we notice that the two parties have different focuses. We therefore use the log-odds ratio to identify the most representative words for each party in Figure~\ref{fig:logodds}.

We find that the Democratic tweets have over-represented words related to media, such as ``telephone'', ``town hall'', and ``facebook'', while a similar cluster for the Republican tweets appear to be related to the White House and its press conferences, such as ``realdonaldtrump'', ``whitehouse'', and ``press''. Additionally, each party has words related to states, cities, and public figures from these places in the US. Meanwhile, the largest category in the Democratic tweets appears to be about the pandemic, such as ``health'', ``response'', ``covid'', ``emergency'', etc. Another cluster including ``disparities'' and ``disproportionately'' also suggest that they discuss issues about social and racial inequalities more. In the Republican case, few words such as ``inittogether'' appears to be directly related to the pandemic. Only the phrases and hashtags for certain region such as ``covidma'' and ``inthistogetherohio'' are detected, indicating much less active narrative regarding the pandemic from the Republicans.

Lastly, both parties have some unique categories; the Democratic tweets has a cluster related to testing, specifically, including words such as ``tested'' and ``positive''. The Republican tweets has a particular cluster about China and the Chinese Communist Party, reflecting the ex-president's narrative against China.

\begin{figure}[h]
    \centering
\includegraphics[width=\textwidth]{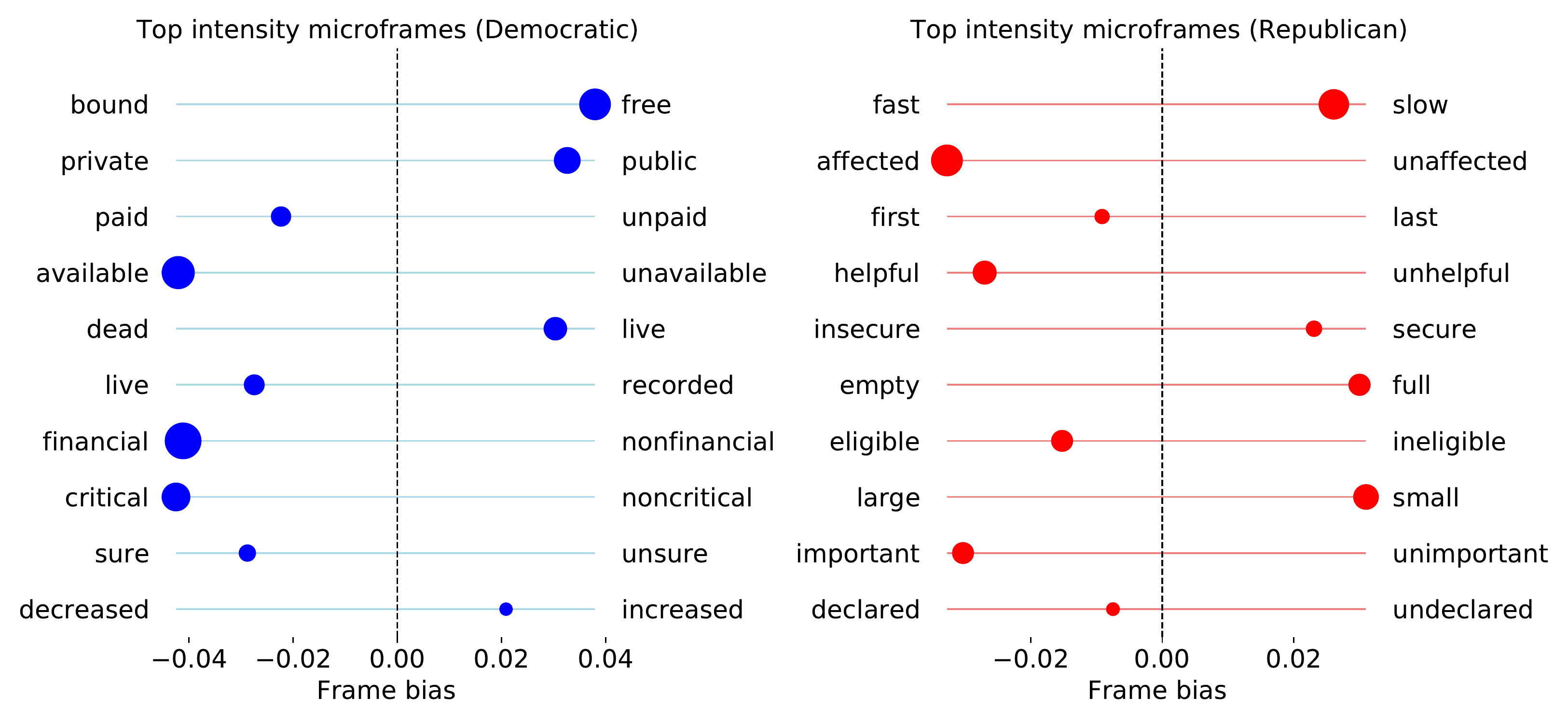}
    \caption{ Top 10 microframes with the largest intensity differences between parties, as well as their frame bias. The position of points indicate the values of bias, and the size of points indicate the values of intensity. The tick labels are the poles of the microframes. }
    \label{fig:frameaxis}
\end{figure}

The overrepresented words give us a sense of the topics and issues that set up the context for each party's narrative frameworks. Our analysis of the framing used in each party's tweets reveals the ways in which they shape their narratives. While the two parties share many common microframes about the pandemic, such as \emph{new versus worn} and \emph{endemic versus epidemic} (see Figure S2), here we focus on the microframes that one party uses significantly more than the other. \Cref{fig:frameaxis} shows the bias and intensity for each of the top ten microframes we identify (see Section~\ref{sec:data}). For example, the Democratic tweets features the \emph{public versus private} frame more intensely than the republican tweets, and at the same time they are more biased towards ``public'' rather than ``private''.

Since it is hard to interpret the pole words without context, we also show the tweets with the highest intensity for each microframe in \Cref{tab:top_tweets}. Combining the pole words and tweet texts, we find that the Democratic frames strongly feature the economic relief during the pandemic, discussing topics such as financial relief, increased funds for support, free testing, etc., which are picked up by the microframe pole words including \emph{free}, \emph{financial}, \emph{increased}, and \emph{paid}.  Additionally, the \emph{public versus private} microframe identifies the emphasis on the public aspect of the pandemic and its response. They also frequently tweet about live events and town hall meetings, invoking the \emph{live} frame. Taken together, we interpret that they emphasize the roles that the government should play regarding the pandemic, contrasting to the Republican framing that we discuss below.

Republican microframes include aid for \emph{small} business, the \emph{eligibility} for financial aid, and \emph{securing} the economy and nation. ``\emph{Slowing} the spread'' appears to be the top slogan used in Republican tweets, emphasizing the roles that individuals play, which contrasts the Democratic narrative. Additionally, the top tweets about \emph{declaring} national emergency, \emph{important} information, and \emph{full} statements also suggests that the Republicans tend to use Twitter as a channel for formal announcements. 

Finally, we examine the characters in each party's narrative frameworks---people who need healthcare, travelers, voters, etc---and their relationships. For insights into how these characters are represented in the politicians' tweets, we explore the semantic roles in these tweets, in particular, the Agents and Patients. We explore the most frequent Agents and Patients in both parties' tweets in Figure S1. We find many common characters as personal pronouns, but also notice some unique semantic roles, such as ``the resources'' and ``lives'' in Democratic tweets, and ``COVID'' and ``relief'' in Republican ones. Furthermore, the Republican tweets often mention the Agent ``Democrats'', and the Democratic tweets often use ``Trump'' and ``the president''.

For a more detailed analysis of the semantic roles, we consider the combinations of an Agent, a verb, and a Patient in each party's tweets. We use the frequency for each combination to identify the most characteristic combinations. We found 321,913 unique combinations in the Democratic tweets and 82,821 unique combinations in Republican tweets. Table \ref{tab:top_combinations} shows the top combinations whose frequency in Democratic tweets is higher than in Republican tweets, and vise versa. 

We find that most of the top combinations from Democratic tweets convey a message of ``they'' need support and ``we'' do everything we can to provide the resources, save lives, etc, further confirming the emphasis on the public response to the pandemic that we found in our framing analysis. Meanwhile, the combinations from Republicans are more diverse, featuring combating COVID, holding press conference, and aiding small businesses. Additionally, one combination discusses the threat of socialism.

\begin{table}
\centering
\begin{tabular}{lll}
Top Democratic combinations   & Top  Republican combinations \\           
\hline
\hline                                                                                                                                                                                                                                                                   they, need, the resources   & we, combat, covid \\
\hline        
 we, can, everything & I, holding, a news conference\\
 \hline        
 we, do, more & covid, impacted, small businesses\\
 \hline        
 they, need, the support & we, fight, covid\\
 \hline        
we, do, everything we can & governor hutchinson, provides, update\\
\hline        
we, save, lives & we, moving, tax day\\
\hline        
I, joined, my colleagues & I, provide, a covid update\\
\hline        
we, do, what & I, holding, a press conference\\
\hline        
those who, need, it & socialism, destroys, nations\\
\hline        
we, recommit, ourselves & covid, affected, those\\
 
\end{tabular}
\caption{Top Agent, verb, and Patient combinations in Democratic and Republican tweets extracted by semantic role labeling with largest differences in frequency. The left column shows the combinations where the frequencies in Democratic tweets are larger than the frequencies in Republican tweets, and vice versa. Most combinations in Democratic tweets focus on resources and support, while combinations in Republican tweets discuss combating COVID, news updates, support for small businesses, and the threat of socialism. }
\label{tab:top_combinations}
\end{table}

From Figure S1, we also notice that the Agents often contains personal pronouns such as ``I'', ``we'', ``they'', and both parties frequently discuss the opposite party, such as the Agent ``Trump'' from Democratic tweets, and ``Democrats'' from Republican tweets, evoking a membership categorization process. We therefore focus on the personal pronouns as Agents that we group into two categories---\emph{us}, including the personal pronouns ``I'', ``we'', ``us'', ``our'', and ``ours'', and \emph{them}, including the words ``they'', ``their'', and ``them''. Additionally, we compile two lists of words associated with ``Democrats'' for Republicans, and vice versa (see Section \ref{sec:data}). 

\begin{figure}[h]
    \centering
\includegraphics[width=0.73\textwidth]{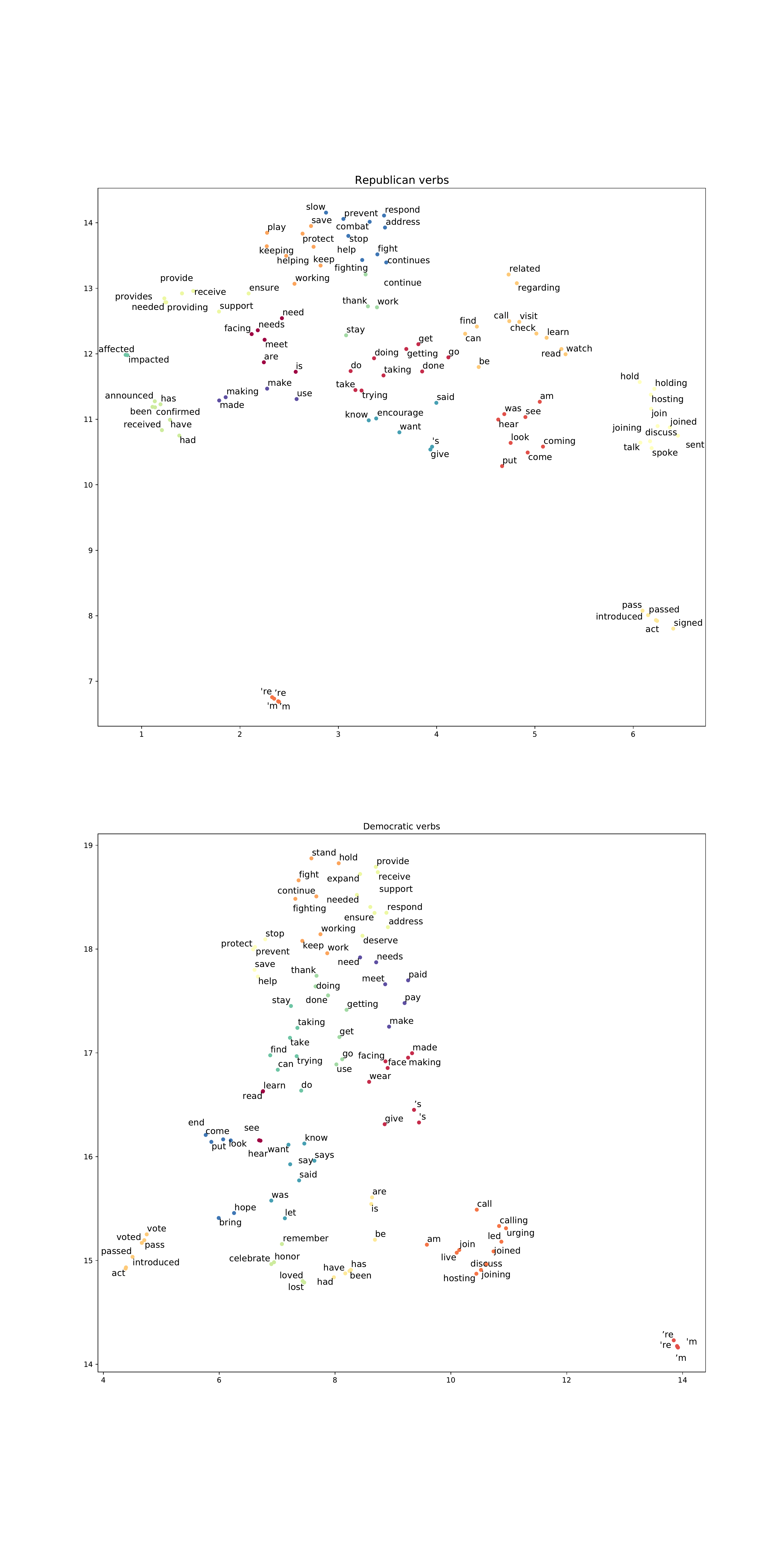}
\vspace{-1in}
    \caption{One-hundred most frequent verbs from Democratic and Republican tweets. Each verb is plotted using their GloVe embeddings with dimensionality reduced to 2 using UMAP. For each party, the verbs are grouped into 15 distinct clusters using the K-means algorithm. Colors of the points indicate cluster membership.}
    \label{fig:verb_clusters}
\end{figure}

We choose specific verbs for a more focused investigation. To leverage the semantic similarities between verbs, we consider the verb clusters that we create from the GloVe embeddings of verbs (see Section~\ref{sec:data} for details). These clusters are shown in Figure~\ref{fig:verb_clusters}. Based on the proximity between verbs and examination of their Patients, we choose three sets of verbs that are most relevant to the pandemic, as well as having a number of diverse semantic roles as their Patients. We then consider the Patients with highest frequency for each set of verbs.

\begin{figure}
    \centering
\begin{subfigure}{\textwidth}
\centering
\includegraphics[width=0.8\textwidth]{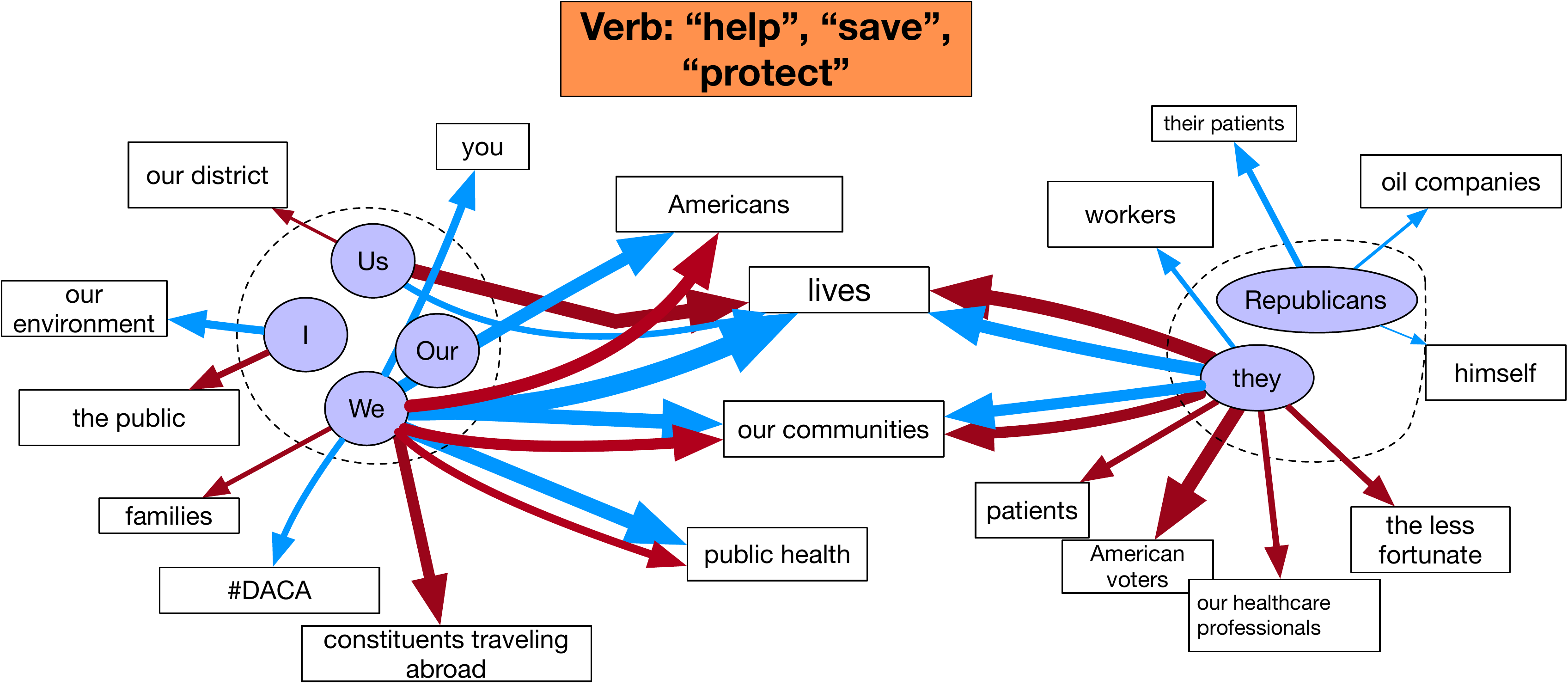}
\end{subfigure}

\begin{subfigure}{\textwidth}
\centering
\includegraphics[width=0.8\textwidth]{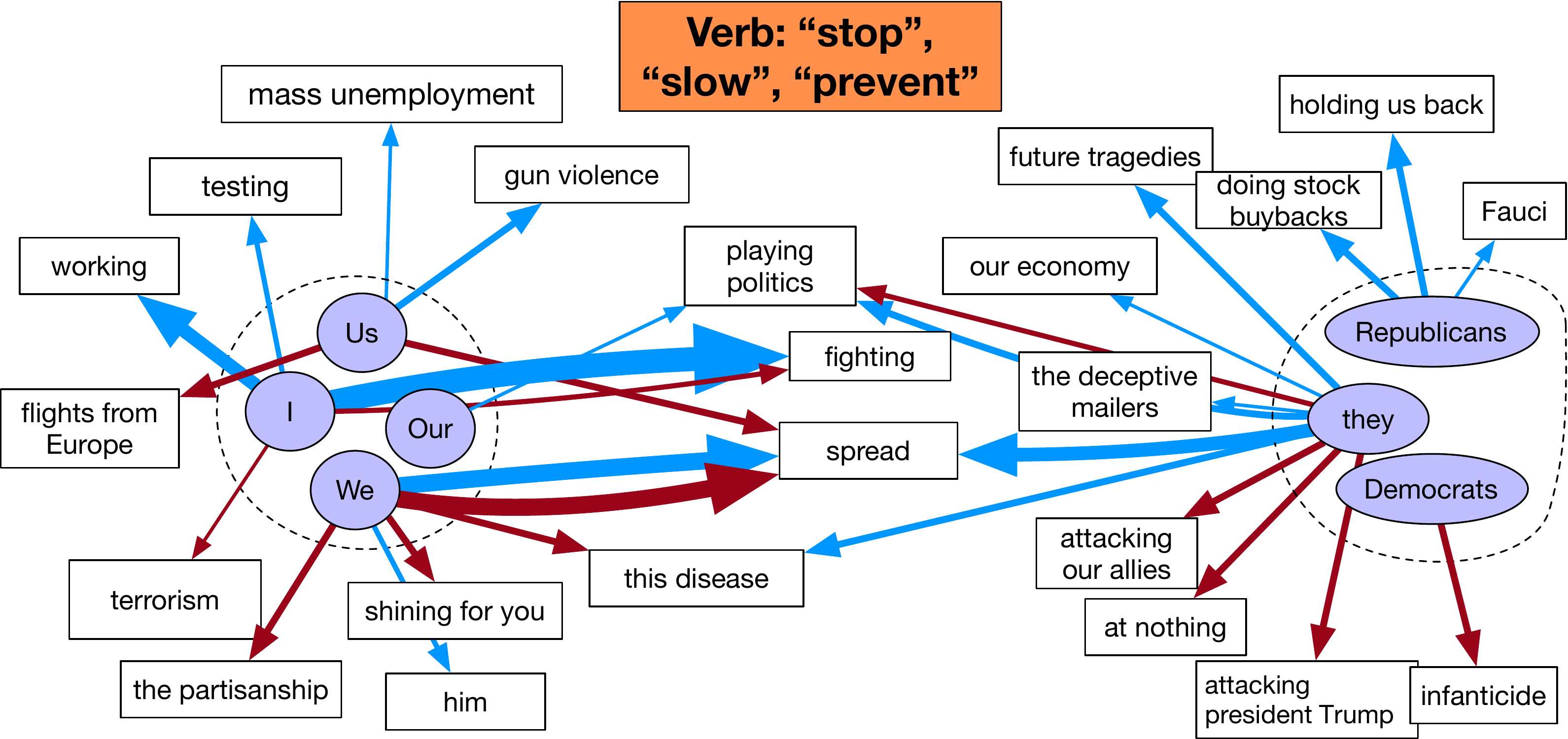}
\end{subfigure}

\begin{subfigure}{\textwidth}
\centering
\includegraphics[width=0.8\textwidth]{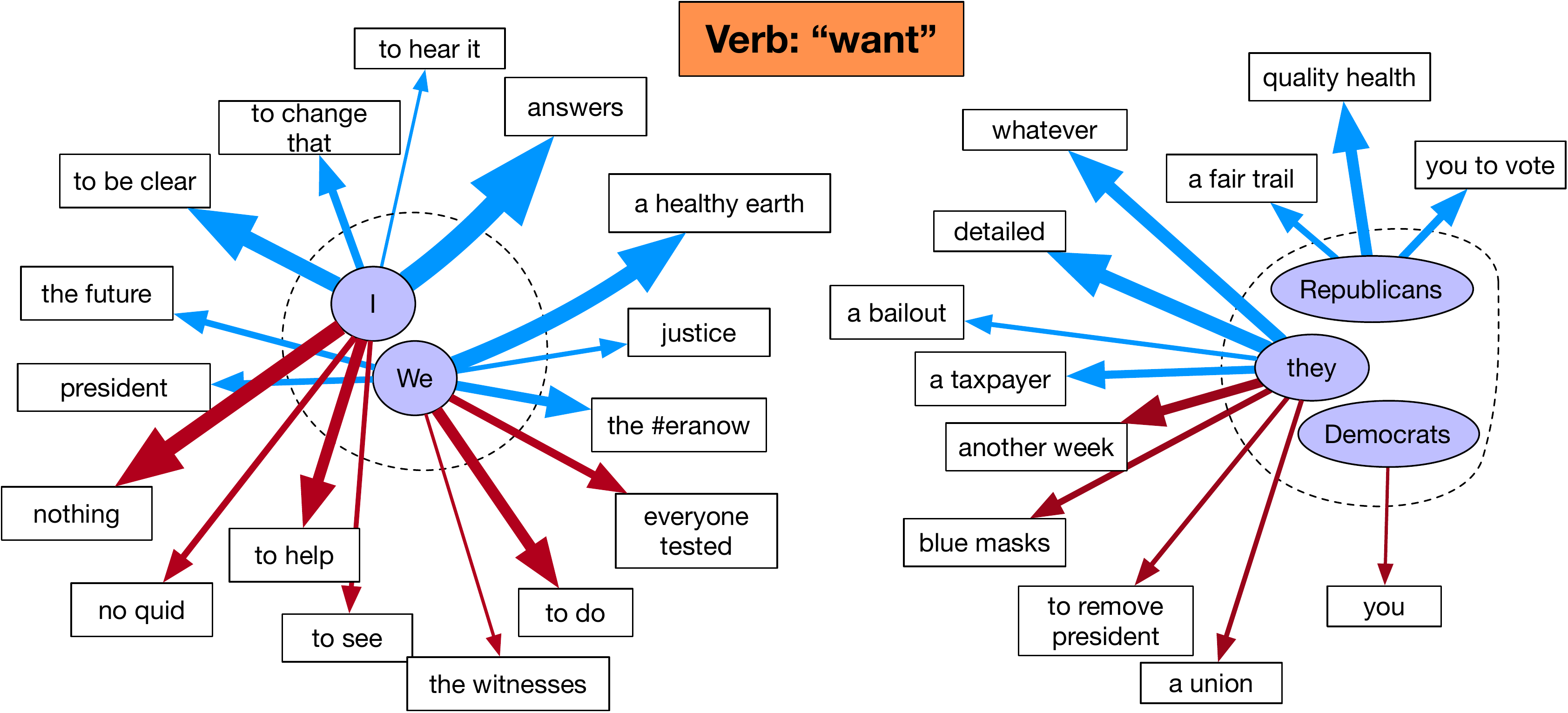}
\end{subfigure}
\caption{Agents and Patients for selected sets of verbs with the highest frequency. Some Patients with similar meanings are combined or omitted. Blue arrows represent relationships found in Democratic tweets, and red in Republican tweets. The sizes of arrows indicate the frequencies of the Patients.}
\label{fig:srl}
\end{figure}

We begin by examining the Patients for the verbs ``help'', ``save'', and ``protect'' in Figure~\ref{fig:srl}. For both of the ``us'' and ``them'' categories, we find a strong shared theme about curbing the pandemic, such as saving lives, helping Americans and public health. Despite some party-specific Patients such as ``\#DACA'' and ``oil companies'', these semantic roles indicate an overlap in both parties' tweets when it comes to protecting American people (although the way they frame help can be different as we discuss above). 

We then move to the set of verbs ``stop'', ``slow'', and ``prevent''. While both parties share a common theme in ``stop the spread'',  we observe many inter-partisan exchanges for both categories. For example, the Democrats discuss stop ``mass employment'' and ``gun violence'',  and the Republicans discuss stop ``terrorism'' as part of their own agendas. In the ``them'' category, the Democrats accuse the Republicans of stopping Fauci and ``doing stock buybacks'', and the Republicans calls for the other party to stop ``attacking president Trump'' and ``the deceptive mailers''. Compared to the previous set, this set of verbs has much less common Patients between two parties.

Finally, we check the verb ``want'' and find that the Patients are rather distinctive for both categories. In the ``us'' category, the Democrats emphasizes ``answers'', ``justice'', ``a healthy earth'', and calling for the Equal Rights Amendment. Meanwhile, the Republicans do not have such strong callings, potentially due to the ruling/opposition party dynamics. In the ``them'' category, we see strong partisan messages about the opposite party, such as the Republican tweets discussing the Democrats' ``blue masks'' and ``to remove president''. This verb does not have any shared Patients, hinting at the different callings from each party.



\section{Discussion}
In this work, we characterize the political narrative frameworks about the COVID-19 crisis constructed by two major U.S. parties, demonstrating that a suite of relatively simple natural language processing methods can be applied to a large dataset to produce useful insights into the diverging narrative frameworks. We examine each narrative framework from three aspects: context, framing, and characters and relationships. We show that the Democratic narrative framework contains more discussion about the pandemic overall, whereas the Republican one includes more mentioning of other political entities. In terms of framing, the Democratic narrative focuses on the financial relief and public health service during the COVID-19 crisis, whereas the Republican narrative emphasizes small business and policy announcements. When we consider the semantic agents, these different focuses are further exposed, and we also found that while both parties find a common ground in battling the pandemic, they also have distinct focuses and political goals, and use their narratives to criticize the other party. 

Our work demonstrates that computational methods can automatically extract strong signatures of political narratives that fit the key theories of political science, providing a useful ``recipe'' for computational text analysis. In addition, we also provide empirical analysis about diverging narrative frameworks in U.S. politics during the pandemic. Our results confirming our intuition, commonsense, and social theories about American politics is a strong evidence for the effectiveness of the tools that we employ. Although we applied the suite of methods to the political communication domain, they are generally applicable to other domains as well. By using an integrated set of computational methods, we bridge the gap between sophisticated NLP methodologies and real-world social  problems. 

Our study has several key limitations. One limitation of our FrameAxis model is not being able to distinguish word senses; for example, it is not able to separate ``live'' as the antonym of ``dead'', and ``live'' as the antonym of ``recorded''. This may lead to confusion when both word senses are widely used in the corpora. tweets with very different topics may also be identified under the same microframe, such as in the case of \emph{available versus unavailable}, where the availability of COVID testing and availability for comment are put together. Such limitations may be partially addressed by using contextualized word embeddings such as ELMO or BERT, and will be an interesting future work.


Our semantic agent analysis use modern SRL tools to automatically identify semantic roles, but the interpretation of such roles remain a challenging task. For example, in Figure~\ref{fig:srl}, manual examination is required to select the Agents and verbs, as well as inferring their context. We are also limited to showing several small sets of verbs and their semantic roles. Additionally, when we examine the membership categorization, some semantic roles such as ``they'' may refer to a third group, instead of one of the parties, and these could not be identified by our model. More automatic ways of analyzing and exploring the SRL data can therefore be fruitful future work.

We analyze the political narratives on a party level by combining all tweets posted by the Democratic or Republican politicians, and do not consider individual politicians. As opinions within a party often diverge, our overarching analysis may miss such dissents. Our study also does not distinguish original tweets and retweets by the politicians.

As we are not working on well-established tasks with systematic benchmarks, and because the tools are exploratory in nature (i.e., they serve as discovery tools and should be combined with human expertise in most cases), it is difficult to quantitatively evaluate them, although we have more rigorous evaluation tasks for our FrameAxis model~\cite{kwak2020frameaxis}. We believe that designing systematic benchmarks for narrative analysis is a challenging, yet important future work.


\section{Abbreviations used in the manuscript}
\noindent \textbf{PDA} \quad \textbf{P}olitical \textbf{D}iscourse \textbf{A}nalysis \\\
\textbf{SRL} \quad \textbf{S}emantic \textbf{R}ole \textbf{L}abeling \\\
\textbf{NLP} \quad \textbf{N}atural \textbf{L}anguage \textbf{P}rocessing \\\
\textbf{UMAP} \quad \textbf{U}niform \textbf{M}anifold \textbf{A}pproximation and \textbf{P}rojection \\\
\textbf{GloVe} \quad \textbf{Gl}obal \textbf{Ve}ctors for Word Representation\\\

\section{Declarations}
\subsection{Availability of data and material}
Our dataset and code will be made available at \url{https://github.com/yzjing/covid19-politics}.

\subsection{Competing interests}
The authors declare that they have no competing interests.

\subsection{Funding}
Not applicable

\subsection{Authors' Contributions}
E.J. and Y.Y.A. designed the study. E.J. collected the data and performed the analysis. E.J. and Y.Y. A wrote the paper.

\subsection{Acknowledgements}
We thank Sandra K\"ubler, Xiaozhong Liu, Minje Kim, Haewoon Kwak, Jisun An, Byungkyu Lee, Matthew Josefy, and the anonymous reviewers for their insightful comments. 

{\scriptsize
\begin{longtable}{p{8cm}p{8cm}}

	\hline\hline
  	Democratic Microframe  & Republican Microframe\\ \hline\hline
		
	bound---free	 & fast---slow\\\hline
	
	 ``Free COVID testing is available near you. '' 
	\newline\newline ``Today's free COVID testing sites''
	\newline\newline ``Testing, testing, testing. the bill makes sure that COVID testing is free for all Americans. ''
	
	& `Do your part to slow the spread of the Wuhan COVID: ''
	\newline\newline ``We all need to do our part to slow the spread of the COVID. here's what you can do to help: ''
	\newline\newline ``rt @housegop: are you doing your part to slow the spread of the COVID?''
	
	\\\hline
	
	decreased---increased & declared---undeclared \\ \hline
	 ``Check to see if you qualify for paid sick leave because of the COVID here''
	\newline\newline  `` Stand with @pattymurray and @sengillibrand and support the paid leave act to provide additional support to workers \& businesses for paid family and sick leave during the COVID outbreak. ''
	\newline\newline ``Today, the house will vote on our next COVID response legislation to provide Americans w\/paid family and medical leave, increased federal medicaid funds to support our state public health partners, free testing, \& emergency sick leave for those impacted by the virus''
	& 
	``My statement after president @realdonaldtrump declared a \#nationalemergency to respond to COVID.'' 
	\newline\newline ``The first public health emergency was declared on March 6 and allows the state to increase coordination across all levels of government in the state's response to COVID.''
	\newline\newline ``President @realdonaldtrump has declared today as national day of prayer. Please join me in praying for our country as we continue to respond to the COVID pandemic.''
	\\ \hline

  	sure---unsure & important---unimportant \\ \hline
	 ``\#MD02 constituents, unsure where to turn for local COVID resources? check out the below graphic for the hotline for your county. ''
	\newline\newline  ``The least the president can do is make sure they have the equipment they need. COVID 3/3''
	\newline\newline ``The response to COVID needs to help all Americans. i'm working with my colleagues to make sure that it does.''
	& 
	``Important information for you and your family about the COVID  '' 
	\newline\newline ``Important information from @cdcgov regarding COVID  ''
	\newline\newline ``Important COVID update from the @deptofdefense in the thread below.''
	\\ \hline

  	critical---noncritical & large---small	\\\hline

	``rt @frankpallone: @WHO is critical in the fight against the COVID pandemic. Trump must work with the world?s premier public health... '' 
	\newline\newline ``rt @uazmedphx: to address the critical needs of the Navajo nation during the COVID outbreak, \#uazmedphx, @repgregstanton, as well as?''
	\newline\newline ``It is critical that we ensure those who have access to any COVID vaccine are not the privileged few, but the many who actually need it most.''
	& 
	``If you own or work for a small business affected by the COVID pandemic, visit my website for information on support for small businesses'' 
	\newline\newline ``Visit learn about the EPCC's grant program for small businesses impacted by the COVID find more helpful EPCC small business resources''
	\newline\newline ``Welcomed news for Georgia small business owners. @sbagov emergency loans are now available to impacted businesses in all 159 counties. COVID''
	\\ \hline

	financial---nonfinancial & eligible---ineligible \\\hline
	
	``Thank you, @abigaildisney, for looking out for the most vulnerable affected by the financial repercussions of COVID. ''
	\newline\newline ``rt @repmalinowski: "the COVID will prey not just on the health of Americans but their financial wellbeing. In its next bill responding...''
    	\newline\newline ``May 1 is quickly approaching, and I know that many marylanders are experiencing severe financial hardship because of the COVID. In this thread you'll find information about financial assistance available in MD.''
	 &
	 ``Alabamians laid off or unpaid due to COVID are eligible for unemployment compensation ''
	\newline\newline ``rt @oronline: if you work in Pennsylvania and the novel COVID has affected your job, you may be eligible for benefits. ''
	\newline\newline ``Small businesses: you may be eligible for up to \$2 million in @sbagov low-interest loans if your business has been affected by the COVID. These loans can help fill your working capital needs. Non-profits may also be eligible. Apply online here:  ''
	\\\hline

	live---recorded & empty---full \\\hline
	
	``Tune in now: I'm hosting a Facebook live town hall with @repbillfoster and @repcasten. We will be answering your questions on COVID. Watch live here: ''
	\newline\newline ``tune in now for my Facebook live COVID town hall with @stevelockhartmd of @sutterhealth:''
	\newline\newline ``I am \#live now on Facebook addressing your questions and concerns about the COVID. Tune in here:  ''

	 &
	  ``Read here: my full statement in support of the COVID relief legislation the House just passed.   '' 
	\newline\newline ``My full statement on presumptive COVID cases in South Dakota ''
	\newline\newline ``See my full statement on president @realdonaldtrump's new actions to fight COVID here''
	\\\hline

	dead---live	 & insecure---secure	\\\hline
	
	 ``Tune in now: I'm hosting a Facebook live town hall with @repbillfoster and @repcasten. We will be answering your questions on COVID. Watch live here:  '' 
	 \newline\newline ``I am \#live now on Facebook addressing your questions and concerns about the COVID. tune in here:  ''
	\newline\newline ``As of 2pm today 1,700 people in my state new jersey are tragically dead from COVID and 16,642 Americans are dead across the country.''
	
	 &  ``rt @waysandmeansgop: in the phase three package to secure our economy as we fight against COVID, @ustreasury secretary @stevenmnuchi? ''
	\newline\newline ``I also thank our brave frontline @tsa officers for the risks they face on our behalf, continuing to keep our nation safe \& secure in the COVID pandemic.''
	\newline\newline ``rt @waysandmeansgop: Dems voted against the phase three package to secure our economy as we fight against COVID. This package include?''
		\\\hline

	available---unavailable & helpful---unhelpful	\\\hline
	
	``More information is available from @cdcgov here: COVIDupdates COVIDUS'' 
	\newline\newline ``Free COVID testing is available near you.''
	\newline\newline ``rt @sfpelosi: 77,000 Americans killed by COVID unavailable for comment.''
	
	& 
	"This is a helpful resource for hoosiers to stay updated on COVID " 
	\newline\newline ``Here's some helpful information on COVID for pregnant women and parents from the @cdcgov. You can find these and other resources on my website at  ''
	\newline\newline ``Continue to follow @cdcgov for the latest updates on the COVID and helpful information. \#MI06 ''
	\\\hline

	paid---unpaid  & first---last\\\hline
	
	``Check to see if you qualify for paid sick leave because of the COVID here '' 
	\newline\newline ``rt @facttank: new: as COVID spreads, which U.S. workers have paid sick leave ? and which don't?  ''
	\newline\newline ``I stand with @pattymurray and @sengillibrand and support the paid leave act to provide additional support to workers \& businesses for paid family and sick leave during the COVID outbreak.''
	& 
	 ``Love this. @starbucks is fueling our first responders on the frontlines of the COVID crisis! \#inittogether ''
	\newline\newline ``'rt @chadsabadie: @repabraham: the first responders, you bring calm to chaos COVID''
	\newline\newline ``rt @woodtv: @rephuizenga pitches COVID aid bill for doctors, nurses and other first responders:''
	\\\hline
	
	private---public	 & affected---unaffected\\\hline
	``rt @indivisibleteam: medicines, like the COVID vaccine, that are developed with public money should benefit public health, not create?'' 
	\newline\newline ``@unitedwaydenver @cohealth coloradans can call the cohelp line for the latest public health information on the COVID at 1-877-462-2911. ''
	\newline\newline ``rt @bryan\_pietsch: healthcare workers battling the COVID would have their public and private student loans forgiven under a new bill?''
				
	 & 
	 ``If you own or work for a small business affected by the COVID pandemic, visit my website for information on support for small businesses''
	\newline\newline ``If you own a small business and your operations are being affected by COVID you may be able to get assistance from @sbagov. More info here:  ''
	\newline\newline ``Appeared on @foxbusiness to discuss congressional action being taken to help Americans affected by COVID''
	\\
	\hline

          \caption{Three top tweets from each microframe with the largest difference in intensity between two parties. URLs, emojis, and some special characters are omitted.}
     	\label{tab:top_tweets}
\end{longtable}}

\printbibliography

\end{document}



\title{Supplementary Material for: Characterizing Partisan Political Narrative Frameworks about COVID-19 on Twitter} 
\date{\today}
\maketitle 

\begin{table}[h]
\small
\begin{center}
\begin{tabular}{p{5cm}p{5cm}p{5cm}}
\hline
 Top Democratic tokens compared to backgound &  Top Republican tokens compared to backgound & 
 Common over-represented tokens between two parties \\
\hline
outbreak, novel, symptoms,
 hospitalized,
 related,
 vaccine,
 essex,
 beshear,
 presumptive,
 teletown
  & 
  novel,
 outbreak,
 msdh,
 wvdhhr,
 azdhs,
 hospitalized,
 recovered,
 plasma,
 hysteria,
preparations
  &
 outbreak,
 novel,
 symptoms,
 hospitalized,
 presumptive,
 recovered,
 plasma,
 shortness'
 spreads,
 slowing
   \\

\hline
\end{tabular}
\caption{Over-represented Democratic and Republican tokens compared to the background corpus, and the common over-represented tokens between two parties, computed by comparing the difference in the dense rank for each token in each corpus.} 
\label{table:tweet_size}
\end{center}
\end{table}

\begin{figure}
  \centering
  \begin{subfigure}{\textwidth}
\centering
\includegraphics[width=\textwidth]{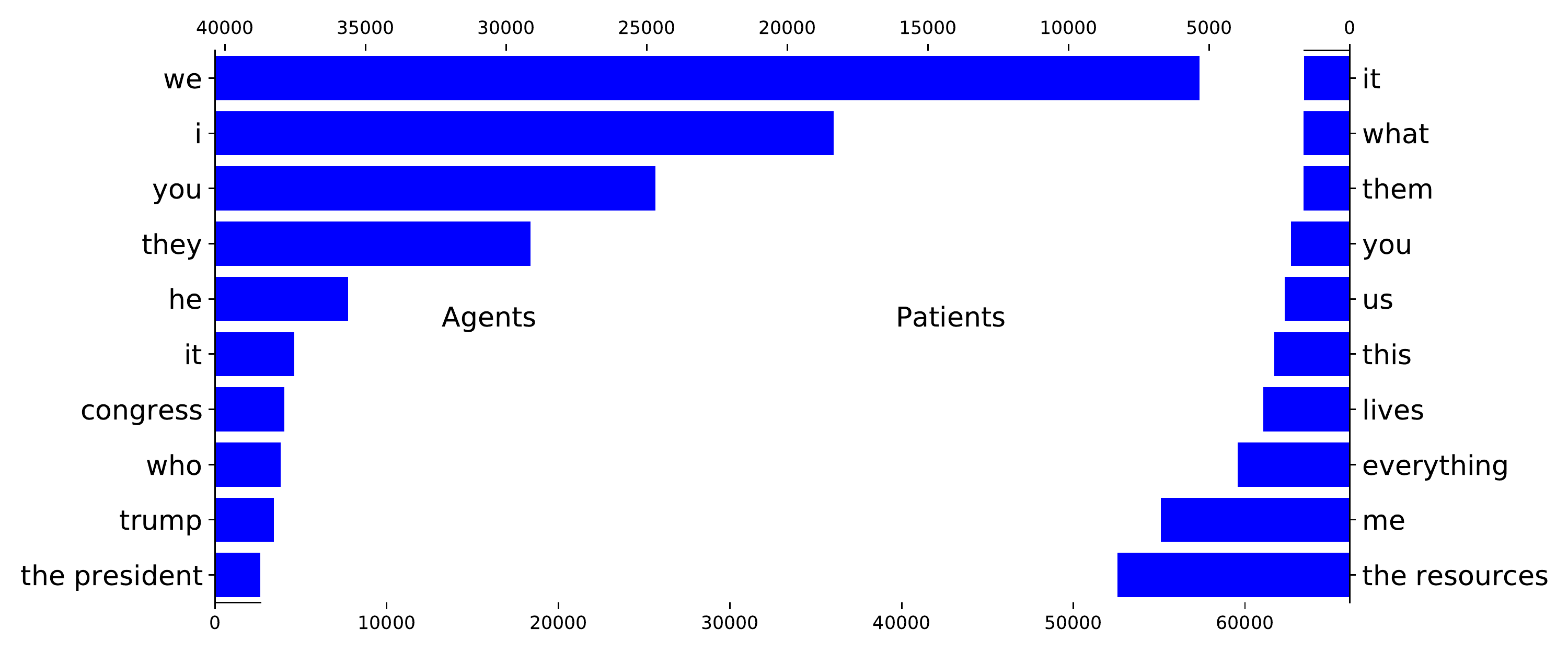}
\end{subfigure}
\begin{subfigure}{\textwidth}
\centering
\includegraphics[width=\textwidth]{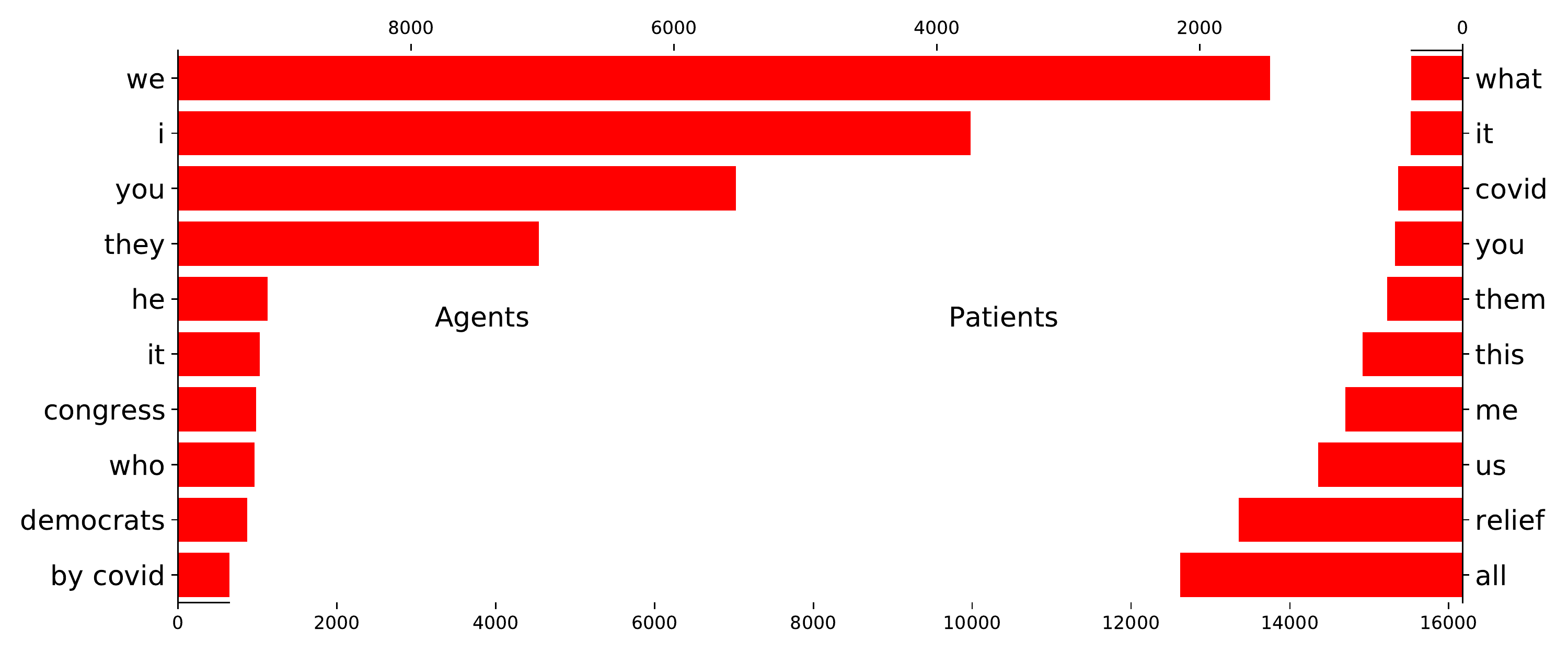}
\end{subfigure}
\caption{Ten most frequent Agents and Patients in Democratic and Republican tweets, with their frequencies. Top figure shows the Democratic Agents and Patients, and bottom figure shows the Republican ones.}
\label{fig:args}
\end{figure}

\begin{figure}[h]
    \centering
\includegraphics[width=\textwidth]{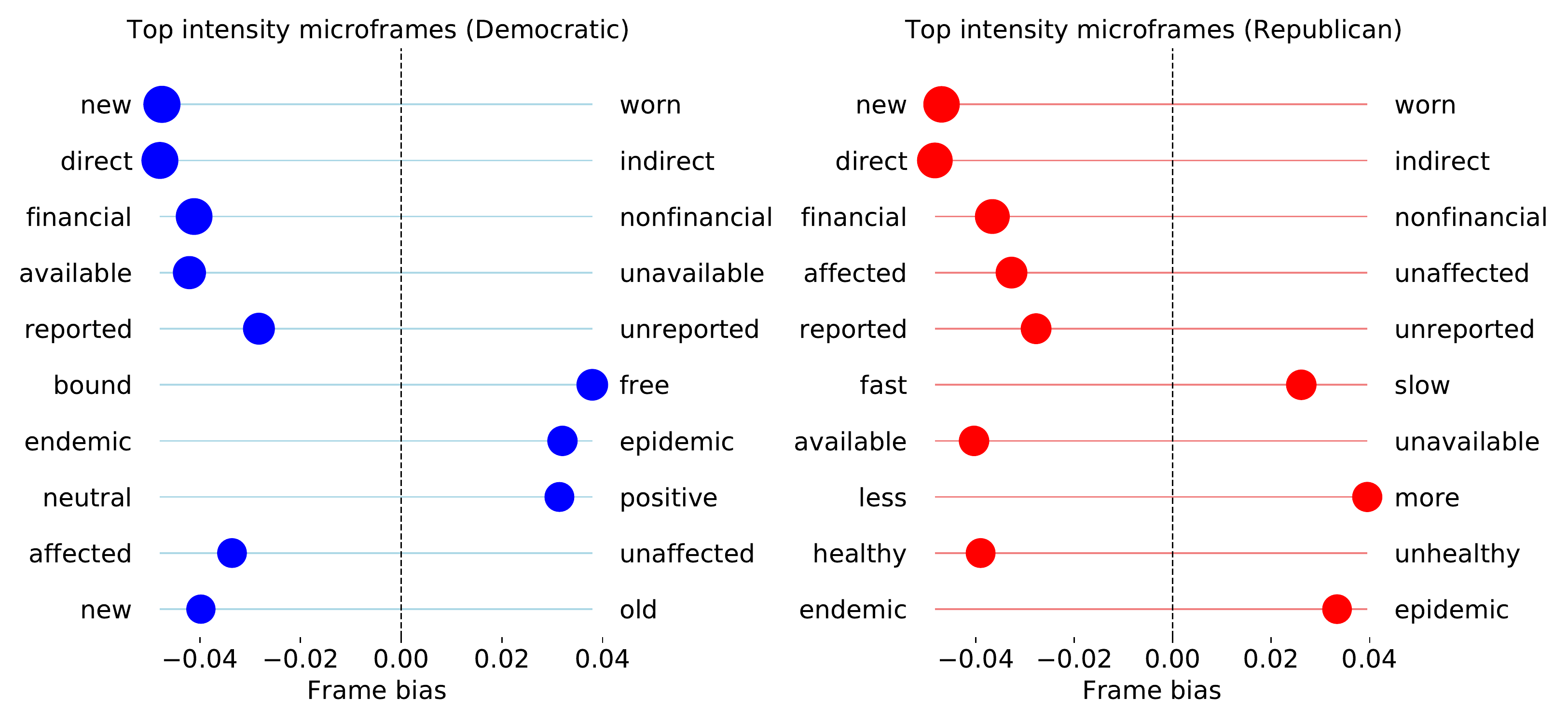}
    \caption{ Top 10 microframes with highest intensity values for each party, as well as their frame bias. The position of points indicate the values of bias, and the size of points indicate the values of intensity. The tick labels are the poles of the microframes. }
    \label{fig:frameaxis}
\end{figure}

\printbibliography{}